%% file: reg.tex
\documentclass[11pt]{article}
\usepackage[top=1in, bottom=1in, left=1in, right=1in]{geometry}
\usepackage{yub}
\usepackage{algorithm,algorithmic}
\usepackage{footnote}
\usepackage{cleveref}
\makesavenoteenv{tabular}

\def\q{\mathsf{q}}

\def\st{\mathsf{SoftThreshold}}
\def\proj{{\rm Proj}}
\def\prox{{\rm prox}}
\def\pq{{\sc ProxQuant}}
\def\signch{{\sf SignChange}}

\title{\pq: Quantized Neural Networks via Proximal Operators}
\author{Yu Bai\footnote{Department of Statistics, Stanford
    University. {\tt yub@stanford.edu}. Work performed at Amazon AI.}
  \and
  Yu-Xiang Wang\footnote{Computer Science Department, UC Santa
    Barbara. {\tt yuxiangw@cs.ucsb.edu}. Work performed at Amazon AI.}
  \and
  Edo Liberty\footnote{Amazon AI. {\tt libertye@amazon.com}.}}

\def\shownotes{0}  
\ifnum\shownotes=1
\newcommand{\authnote}[2]{$\ll$\textsf{\footnotesize #1 notes: #2}$\gg$}
\else
\newcommand{\authnote}[2]{}
\fi

\makeatletter
\def\blfootnote{\gdef\@thefnmark{}\@footnotetext}
\makeatother

\begin{document}
\maketitle

\input{Sections/abstract.tex}
\input{Sections/intro.tex}

\input{Sections/prelim.tex}
\input{Sections/algorithm.tex}
\input{Sections/experiment-new.tex}
\input{Sections/theory.tex}
\input{Sections/conclusion.tex}
\input{Sections/acknowledgement.tex}

\bibliographystyle{abbrvnat}
\bibliography{bib}

\appendix
\input{Sections/wasserstein.tex}
\input{Sections/appendix-experiment.tex}
\input{Sections/signchange.tex}
\input{Sections/proof.tex}

\end{document}

%% file: Sections/abstract.tex
\begin{abstract}

  To make deep neural networks feasible in resource-constrained
  environments (such as mobile devices), it is beneficial to quantize
  models by using low-precision weights.  One common technique for
  quantizing neural networks is the straight-through gradient method,
  which enables back-propagation through the quantization mapping.
  Despite its empirical success, little is understood about why the
  straight-through gradient method works.

  Building upon a novel observation that the straight-through gradient
  method is in fact \emph{identical} to Nesterov's dual-averaging
  algorithm on a quantization constrained optimization problem, we
  propose a more principled alternative approach, called \pq, that
  formulates quantized network training as a regularized learning
  problem instead and optimizes it via the prox-gradient method.
  \pq~does back-propagation on the underlying full-precision vector
  and applies an efficient prox-operator in between stochastic
  gradient steps to encourage quantizedness.  For quantizing ResNets
  and LSTMs, \pq~outperforms state-of-the-art results on binary
  quantization and is on par with state-of-the-art on multi-bit
  quantization.  We further perform theoretical analyses
  showing that \pq~converges to stationary points under mild
  smoothness assumptions, whereas variants such as lazy
  prox-gradient method can fail to converge in the same setting.
  \blfootnote{Code available at~
    \url{https://github.com/allenbai01/ProxQuant}.}

\end{abstract}

%% file: Sections/intro.tex

\section{Introduction}
Deep neural networks (DNNs) have achieved impressive results in
various machine learning
tasks~\citep{GoodfellowBeCo16}. High-performance DNNs typically have
over tens of layers and millions of parameters,
resulting in a high memory usage and a high computational
cost at inference time. However, these networks are often desired in
environments with limited memory and computational power (such as
mobile devices), in which case we would like to compress the network
into a smaller, faster network with comparable performance.

A popular way of achieving such compression is through quantization --
training networks with low-precision weights and/or activation
functions. In a quantized neural network, each weight and/or
activation can be representable in $k$ bits, with a possible codebook
of negligible additional size compared to the network itself. For
example, in a binary neural network ($k=1$), the weights are
restricted to be in $\set{\pm 1}$. Compared with a 32-bit single
precision float, a quantized net reduces the memory usage to $k/32$ of
a full-precision net with the same architecture~\citep{HanMaDa15,
  CourbariauxBeDa15,RastegariOrReFa16, HubaraCoSoElBe17,
  ZhouWuNiZhWeZo16, ZhuHaMaDa16}.
In addition, the
structuredness of the quantized weight matrix can often enable faster
matrix-vector product, thereby also accelerating
inference~\citep{HubaraCoSoElBe17, HanLiMaPuPeHoDa16}.

\begin{figure}[h!]
  \centering
  \begin{subfigure}[b]{0.48\textwidth}
    \includegraphics[width=\textwidth]{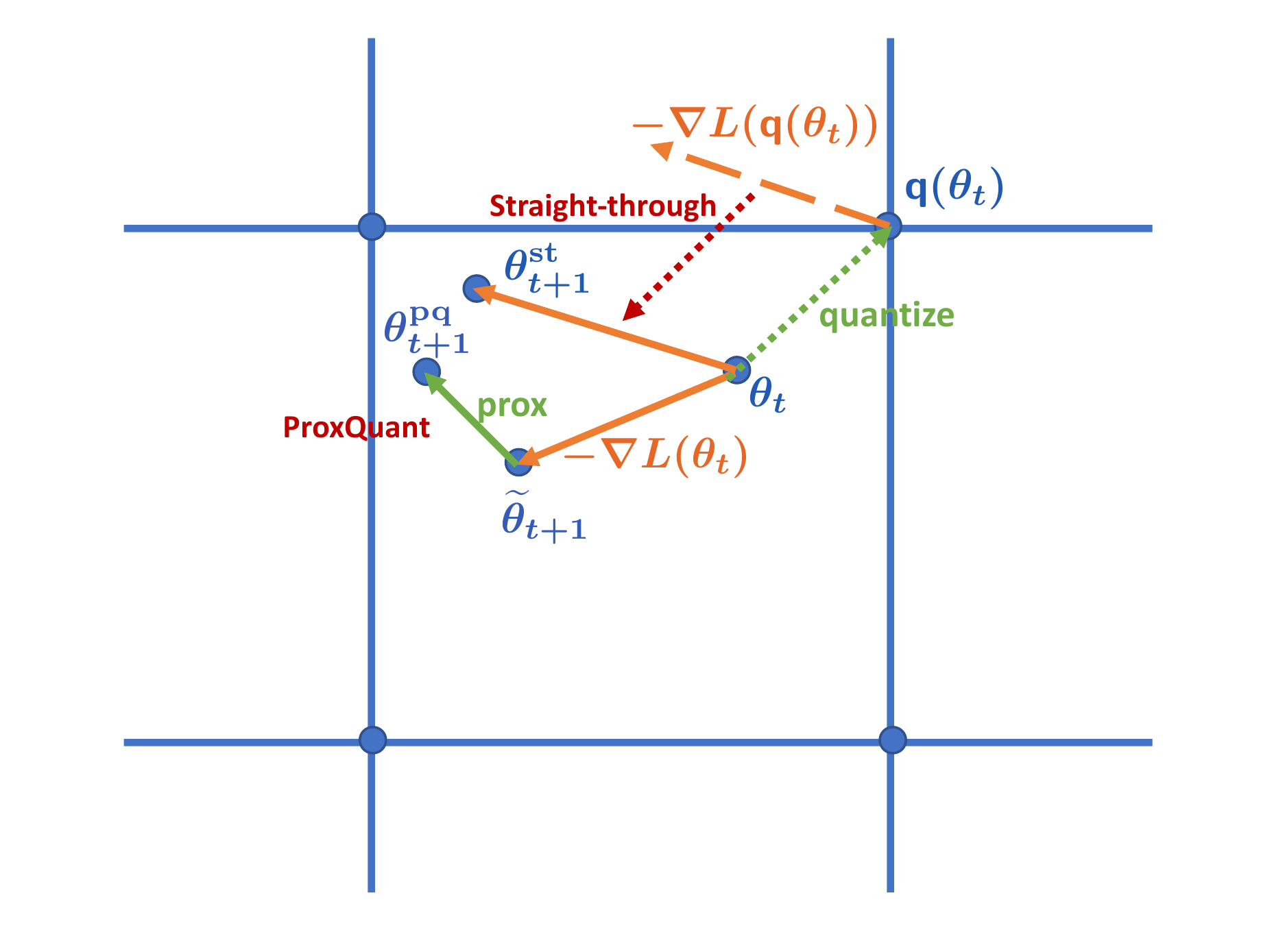}
    \caption{}
    \label{figure:quantization}
  \end{subfigure}
  ~
  \begin{subfigure}[b]{0.48\textwidth}
    \includegraphics[width=\textwidth]{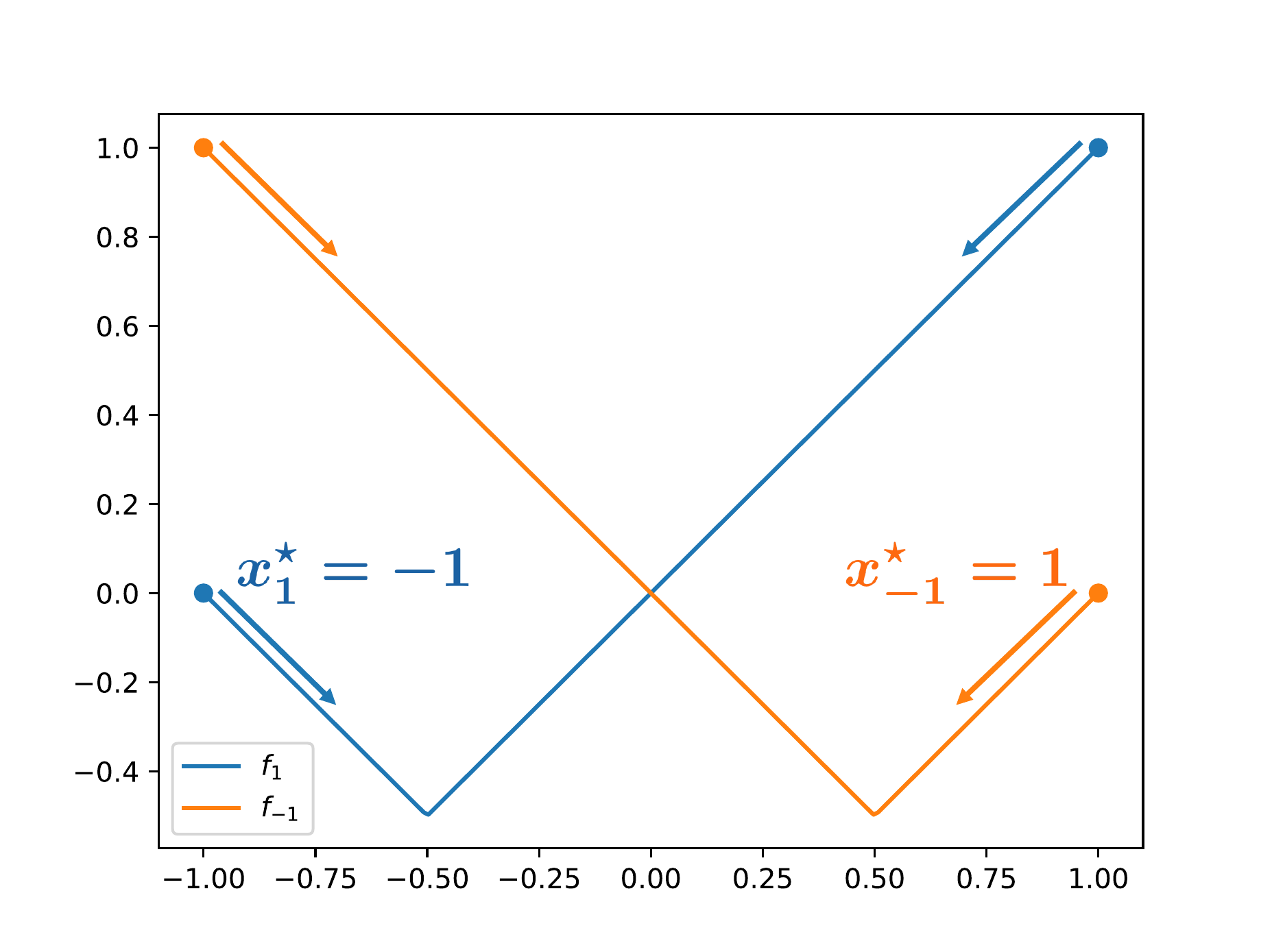}
    \caption{}
    \label{figure:bc-sucks}
  \end{subfigure}
  \caption{\small (a) Comparison of the
    straight-through gradient method and our \pq~method. The
    straight-through method computes the gradient at the quantized
    vector and performs the update at the original real vector;
    \pq~performs a gradient update at the current real vector followed
    by a prox step which encourages quantizedness.
    (b) A two-function toy failure case for
    BinaryConnect. The two functions are $f_1(x)=|x+0.5|-0.5$ (blue)
    and $f_{-1}(x)=|x-0.5|-0.5$ (orange). The derivatives of $f_1$ and
    $f_{-1}$ coincide at $\set{-1,1}$, so any algorithm that only uses
    this information will have identical behaviors on these two
    functions. However, the minimizers in $\set{\pm 1}$ are
    $x_1^\star=-1$ and $x_{-1}^\star=1$, so the algorithm must fail
    on one of them.}
  \label{figure:figure}
\end{figure}

Typically, training a quantized network involves (1) the design of a
\emph{quantizer} $\q$ that maps a full-precision parameter to a
$k$-bit quantized parameter, and (2) the 
\emph{straight-through gradient method}~\citep{CourbariauxBeDa15} that
enables back-propagation from the quantized parameter back onto the
original full-precision parameter, which is critical to the success of
quantized network training.
With quantizer $\q$, an iterate of the straight-through
gradient method (see Figure~\ref{figure:quantization}) proceeds as
$\theta_{t+1}=\theta_t - \eta_t \wt{\grad}L(\theta)|_{\theta=\q(\theta_t)}$, and
$\q(\what{\theta})$ (for the converged $\what{\theta}$) is taken as
the output model. For training binary networks, choosing
$\q(\cdot)=\sign(\cdot)$ gives the BinaryConnect
method~\citep{CourbariauxBeDa15}.

Though appealingly simple and empirically effective, it is
information-theoretically rather mysterious why the straight-through
gradient method works
well, at least in the binary case: while the
goal is to find a parameter $\theta\in\set{\pm 1}^d$ with low loss,
the algorithm only has access to stochastic gradients at
$\set{\pm 1}^d$. As this is a discrete set, {\it a priori}, gradients
in this set do not necessarily contain any information about the function values.
Indeed, a simple one-dimensional example
(Figure~\ref{figure:bc-sucks}) shows that BinaryConnect fails to find
the minimizer of fairly simple convex Lipschitz functions in $\set{\pm
  1}$, due to a lack of gradient information in between.

In this paper, we formulate the problem of model quantization as a
regularized learning problem and propose to solve it with a proximal
gradient method. Our contributions are summarized as follows.
\begin{itemize}
\item We present a unified framework for defining regularization
  functionals that encourage binary, ternary, and multi-bit quantized
  parameters, through penalizing the distance to quantized sets (see
  Section~\ref{section:wasserstein-reg}). For binary quantization, the
  resulting regularizer is a $W$-shaped non-smooth regularizer, which
  shrinks parameters towards either $-1$ or $1$ in the same way that
  the $L_1$ norm regularization shrinks parameters towards $0$.
\item We propose training quantized networks using
  \pq~(Algorithm~\ref{algorithm:proxquant}) --- a stochastic proximal
  gradient method with a homotopy scheme. Compared with the
  straight-through gradient method, \pq~has access to additional
  gradient information at non-quantized points, which avoids the
  problem in Figure~\ref{figure:bc-sucks} and its homotopy scheme
  prevents potential overshoot early in the training
  (Section~\ref{section:homotopy}).

\item We demonstrate the effectiveness and flexibility of \pq~through
  systematic experiments on (1) image classification with
  ResNets (Section~\ref{section:cifar-10}); (2)
  language modeling with LSTMs (Section~\ref{section:lstm}). The
  \pq~method outperforms the state-of-the-art results on binary
  quantization and is comparable with the state-of-the-art on
  ternary and multi-bit quantization.
\item We perform a systematic theoretical study of quantization
  algorithms, showing that our \pq~(standard prox-gradient method)
  converges to stataionary points under mild smoothness assumptions
  (Section~\ref{section:convergence-pq}), where as lazy prox-gradient
  method such as BinaryRelax~\citep{YinZhLyOsQiXi18} fails to converge
  in general (Section~\ref{section:nonconvergence-lazy}). Further, we
  show that BinaryConnect has a very stringent condition to converge
  to any fixed point (Section~\ref{section:convergence-bc}), which we
  verify through a sign change experiment
  (Appendix~\ref{appendix:sign-change}).
\end{itemize}

\subsection{Prior work}
\paragraph{Methodologies}
\citet{HanMaDa15} propose Deep Compression, which compresses a DNN via
sparsification, nearest-neighbor clustering, and Huffman coding. This
architecture is then made into a specially designed hardware for
efficient inference~\citep{HanLiMaPuPeHoDa16}. In a parallel line of
work, \citet{CourbariauxBeDa15} propose BinaryConnect that enables the
training of binary neural networks, and~\citet{LiLi16,ZhuHaMaDa16}
extend this method into ternary quantization. Training and inference
on quantized nets can be made more efficient by also quantizing the
activation~\citep{HubaraCoSoElBe17,RastegariOrReFa16,ZhouWuNiZhWeZo16},
and such networks have achieved impressive performance on large-scale
tasks such as ImageNet
classification~\citep{RastegariOrReFa16,ZhuHaMaDa16} and object
detection~\citep{YinZhQiXi16}. In the NLP land, quantized language
models have been successfully trained using alternating multi-bit
quantization~\citep{XuYaLiOuCaWaZh18}.

\paragraph{Theories} \citet{LiDeXuStSaGo17} prove the convergence
rate of stochastic rounding and BinaryConnect on convex problems and
demonstrate the advantage of BinaryConnect over stochastic rounding on
non-convex problems.  \citet{AndersonBe17} demonstrate the
effectiveness of binary networks through the observation that the
angles between high-dimensional vectors are approximately preserved
when binarized, and thus high-quality feature extraction with binary
weights is possible.~\citet{DingLiSh18} show a universal approximation
theorem for quantized ReLU networks.

\paragraph{Principled methods}
\citet{Sun18} perform model quantization through a Wasserstein
regularization term and minimize via the adversarial representation,
similar as in Wasserstein GANs~\citep{ArjovskyChBo17}. Their method has
the potential of generalizing to other generic requirements on the
parameter, but might be hard to tune due to the instability of the
inner maximization problem.

Prior to our work, a couple of proximal or regularization based
quantization algorithms were proposed as alternatives to the
straight-through gradient method, which we now briefly review and
compare with.  \citep{YinZhLyOsQiXi18} propose BinaryRelax, which
corresponds to a lazy proximal gradient
descent. \citep{HouYaKw17,HouKw18} propose a proximal Newton method
with a diagonal approximate Hessian.  \citet{Carreira17,CarreiraId17}
formulate quantized network training as a constrained optimization
problem and propose to solve them via augmented Lagrangian
methods. Our algorithm is different with all the aformentioned work in
using the non-lazy and ``soft'' proximal gradient descent with a
choice of either $\ell_1$ or $\ell_2$ regularization, whose advantage
over lazy prox-gradient methods is demonstrated both theoretically
(Section~\ref{section:theory}) and experimentally
(Section~\ref{section:cifar-10} and Appendix~\ref{appendix:sign-change}).

%% file: Sections/prelim.tex
\section{Preliminaries}
The optimization difficulty of training quantized models is that they
involve a discrete parameter space and hence efficient local-search
methods are often prohibitive. For example, the problem of training a
binary neural network is to minimize $L(\theta)$ for
$\theta\in\set{\pm 1}^d$. Projected SGD on this set will not move
unless with an unreasonably large stepsize~\citep{LiDeXuStSaGo17},
whereas greedy nearest-neighbor search requires $d$ forward passes
which is intractable for neural networks where $d$ is on the order of
millions. Alternatively, quantized training can also be cast as
minimizing $L(\q(\theta))$ for $\theta\in\R^d$ and an appropriate
\emph{quantizer} $\q$ that maps a real vector to a nearby quantized
vector, but $\theta\mapsto \q(\theta)$ is often non-differentiable and
piecewise constant (such as the binary case $\q(\cdot)=\sign(\cdot)$),
and thus back-propagation through $\q$ does not work.

\subsection{The straight-through gradient method}
\label{section:straight-through}
The pioneering work of BinaryConnect~\citep{CourbariauxBeDa15} proposes
to solve this problem via the \emph{straight-through gradient
  method}, that is, propagate the gradient with respect to
$\q(\theta)$ unaltered to $\theta$, i.e. to let
$\frac{\partial L}{\partial \theta} \defeq \frac{\partial L}{\partial
  \q(\theta)}$. One iterate of the straight-through gradient method
(with the SGD optimizer) is
\begin{equation*}
  \theta_{t+1} = \theta_t - \eta_t \wt{\grad}L(\theta)|_{\theta=\q(\theta_t)}.
\end{equation*}
This enables the real vector $\theta$ to move in the entire Euclidean
space, and taking $\q(\theta)$ at the end of training gives a valid
quantized model. Such a customized back-propagation rule yields good
empirical performance in training quantized nets and has thus become a
standard practice~\citep{CourbariauxBeDa15,ZhuHaMaDa16,
  XuYaLiOuCaWaZh18}. However, as we have discussed, it is information
theoretically unclear how the straight-through method works, and it
does fail on very simple convex Lipschitz functions
(Figure~\ref{figure:bc-sucks}).


\subsection{Straight-through gradient as lazy projection}
\label{section:lazy-projection}
Our first observation is that the straight-through gradient method is
equivalent to a \emph{dual-averaging} method, or a lazy projected
SGD~\citep{Xiao10}. In the binary case, we wish to minimize
$L(\theta)$ over $\mc{Q}=\set{\pm 1}^d$, and the lazy projected SGD
proceeds as
\begin{equation}
  \left\{
  \begin{aligned}
    & \wt{\theta}_{t} = \proj_{\mc{Q}}(\theta_t) = \sign(\theta_t) =
    \q(\theta_t), \\
    & \theta_{t+1} = \theta_t - \eta_t \wt{\grad} L(\wt{\theta}_t).
  \end{aligned}
  \right.
\end{equation}
Written compactly, this is
$\theta_{t+1}=\theta_t - \eta_t \wt{\grad}
L(\theta)|_{\theta=\q(\theta_t)}$, which is exactly the straight-through
gradient method: take the gradient at the quantized vector and
perform the update on the original real vector.

\subsection{Projection as a limiting proximal operator}
\label{section:projection-as-prox}
We take a broader point of view that a projection is also a limiting
proximal operator with a suitable regularizer, to allow more
generality and to motivate our proposed algorithm. Given any set
$\mc{Q}$, one could identify a regularizer $R:\R^d\to\R_{\ge 0}$ such
that the following hold:
\begin{equation}
  \label{equation:reg-requirement}
  R(\theta)=0,~~\forall \theta\in\mc{Q}~~~{\rm
    and}~~~R(\theta)>0,~~\forall \theta\notin\mc{Q}. 
\end{equation}
In the case $\mc{Q}=\set{\pm 1}^d$ for example, one could take
\begin{equation}
  \label{equation:binary-reg}
  R(\theta) = R_{\rm bin}(\theta) = \sum_{j=1}^d
  \min\set{|\theta_j-1|, |\theta_j+1|}. 
\end{equation}
The proximal operator (or prox operator)~\citep{ParikhBo14} with
respect to $R$ and strength $\lambda>0$ is
\begin{equation*}
  \prox_{\lambda R}(\theta) \defeq \argmin_{\wt{\theta}\in\R^d} \left\{
    \frac{1}{2}\ltwo{\wt{\theta}-\theta}^2 + \lambda R(\wt{\theta}) \right\}.
\end{equation*}
In the limiting case $\lambda=\infty$, the argmin has to satisfy
$R(\theta)=0$, i.e. $\theta\in\mc{Q}$, and the prox operator is to
minimize $\ltwo{\theta-\theta_0}^2$ over $\theta\in\mc{Q}$, which is
the Euclidean projection onto $\mc{Q}$. Hence, projection is also a
prox operator with $\lambda=\infty$, and the straight-through gradient
estimate is equivalent to a lazy proximal gradient descent with and
$\lambda=\infty$.

While the prox operator with $\lambda=\infty$ correponds to ``hard''
projection onto the discrete set $\mc{Q}$, when $\lambda<\infty$ it
becomes a ``soft'' projection that moves towards $\mc{Q}$. Compared
with the hard projection, a finite $\lambda$ is less aggressive and
has the potential advantage of avoiding overshoot early in training.
Further, as the prox operator does not strictly enforce quantizedness,
it is in principle able to query the gradients at every point in the
space, and therefore has access to more information than the
straight-through gradient method.

%% file: Sections/algorithm.tex

\section{Quantized net training via regularized learning}
\label{section:algorithm}
We propose the \pq~algorithm, which adds a quantization-inducing
regularizer onto the loss and optimizes via the (non-lazy)
prox-gradient method with a finite $\lambda$. 
The prototypical version of \pq~is described in
Algorithm~\ref{algorithm:proxquant}.
\begin{algorithm}[H]
  \caption{\pq: Prox-gradient method for quantized net
    training}
  \label{algorithm:proxquant}
  \begin{algorithmic}
    \REQUIRE Regularizer $R$ that induces desired
    quantizedness, initialization $\theta_0$, learning rates
    $\set{\eta_t}_{t\ge 0}$, regularization strengths
    $\set{\lambda_t}_{t\ge 0}$
    \WHILE{not converged}
    \STATE Perform the prox-gradient step
    \begin{align}
      \theta_{t+1}
      & = \prox_{\eta_t\lambda_t R}\left( \theta_t -
        \eta_t\wt{\grad}L(\theta_t) \right). \label{equation:prox-gradient}
    \end{align}
    The inner SGD step in~\cref{equation:prox-gradient} can be
    replaced by any preferred stochastic optimization method such as
    Momentum SGD or Adam~\citep{KingmaBa14}.
    \ENDWHILE
  \end{algorithmic}
\end{algorithm}
Compared to usual full-precision training, \pq~only adds a prox step
after each stochastic gradient step, hence can be implemented
straightforwardly upon existing full-precision training.  As the prox
step does not need to know how the gradient step is performed, our
method adapts to other stochastic optimizers as well such as
Adam.


In the remainder of this section, we define a flexible class of
quantization-inducing regularizers through ``distance to the quantized
set'', derive efficient algorithms of their corresponding prox
operator, and propose a homotopy method for choosing the
regularization strengths.
Our regularization perspective subsumes most existing algorithms for
model-quantization
(e.g.,\citep{CourbariauxBeDa15,HanMaDa15,XuYaLiOuCaWaZh18}) as limits
of certain regularizers with strength $\lambda \rightarrow
\infty$. Our proposed method can be viewed as a principled
generalization of these methods to $\lambda <\infty$ with a non-lazy
prox operator.


\subsection{Regularization for model quantization}
\label{section:wasserstein-reg}
Let $\mc{Q}\subset \R^d$ be a set of quantized parameter vectors.
An ideal regularizer for quantization would be to vanish on
$\mc{Q}$ and reflect some type of distance to
$\mc{Q}$ when $\theta\notin\mc{Q}$. To achieve this, we propose
$L_1$ and $L_2$ regularizers of the form
\begin{equation}
  \label{equation:reg}
  R(\theta) = \inf_{\theta_0\in\mc{Q}}~\lone{\theta - \theta_0}~~~{\rm
    or}~~~R(\theta) = \inf_{\theta_0\in\mc{Q}}~\ltwo{\theta -
    \theta_0}^2.
\end{equation}
This is a highly flexible framework for designing regularizers, as one
could specify any $\mc{Q}$ and choose between $L_1$ and $L_2$.
Specifically, $\mc{Q}$ encodes certain desired quantization structure.
By appropriately choosing $\mc{Q}$, we can specify which part of the
parameter vector to quantize\footnote{Empirically, it is advantageous
  to keep the biases of each layers and the BatchNorm layers at
  full-precision, which is often a negligible fraction, say
  $1/\sqrt{d}$ of the total number of parameters}, the number of bits
to quantize to, whether we allow adaptively-chosen quantization levels
and so on. The choice between \{$L_1$, $L_2$\} will encourage
\{``hard'',``soft''\} quantization respectively, similar as in
standard regularized learning~\citep{Tibshirani96}.


In the following, we present a few examples of regularizers under our
framework \cref{equation:reg} which induce binary weights, ternary
weights and multi-bit quantization.
We will also derive efficient algorithms (or approximation heuristics)
for solving the prox operators corresponding to these regularizers,
which generalize the projection operators used in the straight-through
gradient algorithms.

\begin{wrapfigure}{L}{0.3\textwidth}
  \centering
  \includegraphics[width=0.28\textwidth]{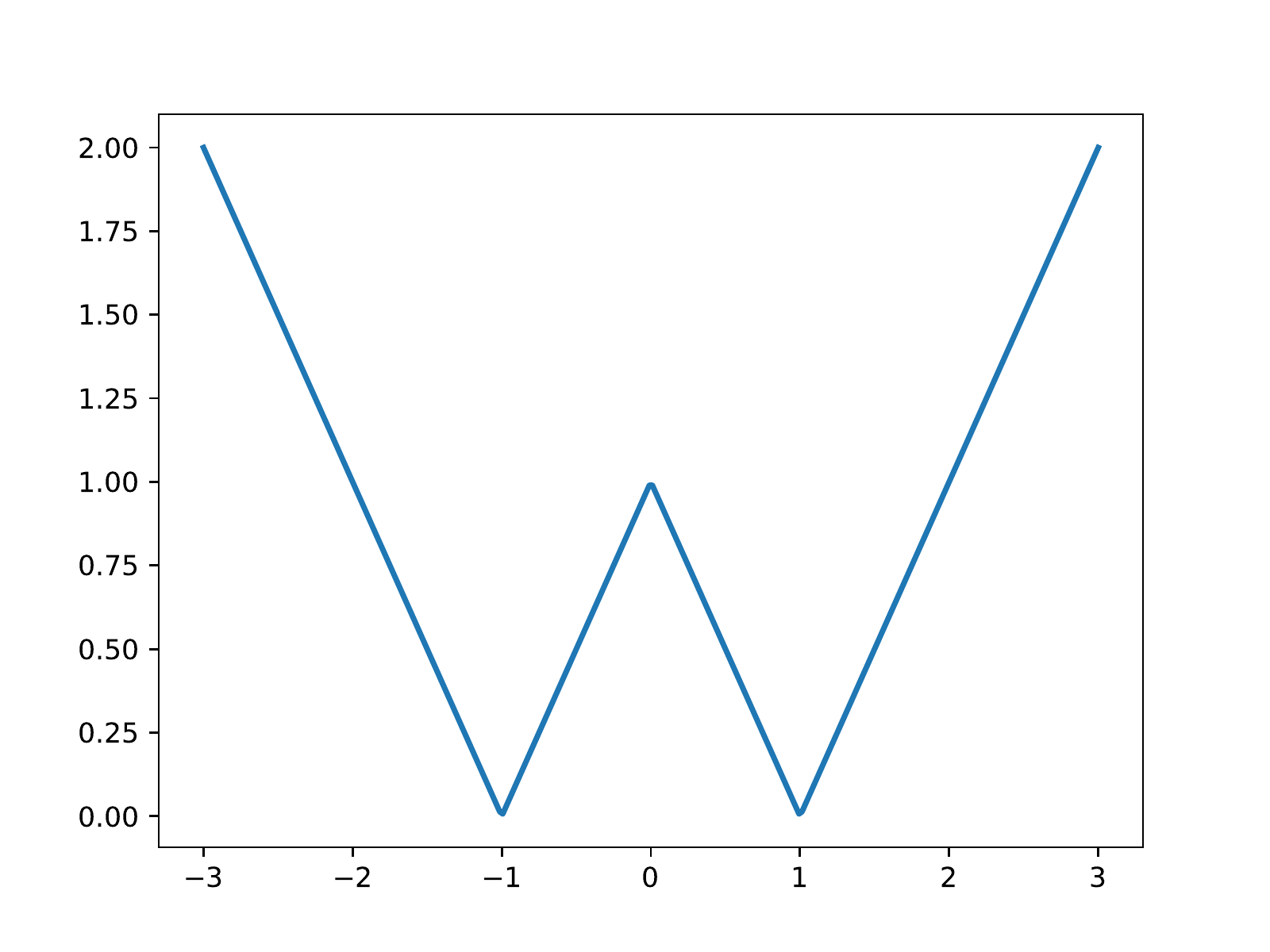}
  \caption{\small W-shaped regularizer for binary quantization.}
  \label{figure:w-reg}
\end{wrapfigure}

\paragraph{Binary neural nets}
In a binary neural net, the entries of $\theta$ are in $\set{\pm
  1}$. A natural choice would be taking $\mc{Q}=\{-1,1\}^d$. The
resulting $L_1$ regularizer is
\begin{equation}
  \label{equation:w-binary}
  \begin{aligned}
    & \quad R(\theta) = \inf_{\theta_0\in\set{\pm 1}^d} \lone{\theta - \theta_0} =  \sum_{j=1}^d \inf_{[\theta_0]_j\in\set{\pm 1}}|\theta_j - [\theta_0]_j| \\
    & = \sum_{j=1}^d \min\set{|\theta_j - 1|, |\theta_j + 1|} = \lone{\theta - \sign(\theta)}.
  \end{aligned}
\end{equation}
This is exactly the binary regularizer $R_{\rm bin}$ that we discussed
earlier in~\cref{equation:binary-reg}.  Figure~\ref{figure:w-reg}
plots the W-shaped one-dimensional component of $R_{\rm bin}$~from
which we see its effect for inducing $\set{\pm 1}$ quantization in
analog to $L_1$ regularization for inducing exact sparsity.

The prox operator with respect to $R_{\rm bin}$, despite being a
non-convex optimization problem, admits a simple analytical solution:
\begin{equation}
  \label{equation:binary-prox}
  \begin{aligned}
    \prox_{\lambda R_{\rm bin}}(\theta)
    & = \st(\theta, \sign(\theta), \lambda) \\
    &  = \sign(\theta) + \sign(\theta - \sign(\theta)) \odot [|\theta -
    \sign(\theta)| - \lambda]_+. 
  \end{aligned}
\end{equation}
We note that the choice of the $L_1$ version is not unique: the
squared $L_2$ version works as well, whose prox operator is given
by~$(\theta+\lambda\sign(\theta))/(1+\lambda)$.  See
Appendix~\ref{appendix:prox} for the derivation of these prox
operators and the definition of the soft thresholding operator.


\paragraph{Multi-bit quantization with adaptive levels.}
Following~\citep{XuYaLiOuCaWaZh18}, we
consider $k$-bit quantized parameters with a structured adaptively-chosen set of quantization levels, which translates into
\begin{equation}
  \label{equation:multi-bit-quant}
  \mc{Q} = \set{ \sum_{i=1}^k \alpha_i b_i:
    \set{\alpha_1,\dots,\alpha_k}\subset\R,~b_i\in\set{\pm 1}^d}
  = \set{\theta_0=B\alpha: \alpha\in\R^k,~B\in\set{\pm 1}^{d\times k}}.
\end{equation}
The squared $L_2$ regularizer for this structure is
\begin{equation}
  \label{equation:multi-bit-w-reg}
  R_{k-{\rm bit}}(\theta) = \inf_{\alpha\in\R^k,B\in\set{\pm
      1}^{d\times k}} \ltwo{\theta - B\alpha}^2,
\end{equation}
which is also the alternating minimization objective
in~\citep{XuYaLiOuCaWaZh18}.


We now derive the prox operator for the
regularizer~\cref{equation:multi-bit-w-reg}. For any $\theta$, we have
\begin{equation}
  \label{equation:multi-bit-prox}
  \begin{aligned}
    & \quad \prox_{\lambda R_{k-{\rm bit}}}(\theta) = \argmin_{\wt{\theta}} \set{\frac{1}{2}\ltwo{\wt{\theta} - \theta}^2 + \lambda\inf_{\alpha\in\R^k,B\in\set{\pm 1}^{d\times k}} \ltwo{\wt{\theta} - B\alpha}^2} \\
    & = \argmin_{\wt{\theta}} \inf_{\alpha\in\R^k,B\in\set{\pm 1}^{d\times k}} \set{\frac{1}{2}\ltwo{\wt{\theta} - \theta}^2 + \lambda \ltwo{\wt{\theta} - B\alpha}^2}.
  \end{aligned}
\end{equation}
This is a joint minimization problem in $(\wt{\theta},B,\alpha)$, and we
adopt an alternating minimization schedule to solve it:
\begin{enumerate}[(1)]
\item Minimize over $\wt{\theta}$ given $(B,\alpha)$, which has a
  closed-form solution
  $\wt{\theta} = \frac{\theta + 2\lambda B\alpha}{1+2\lambda}$.
\item Minimize over $(B,\alpha)$ given $\wt{\theta}$, which does not
  depend on $\theta_0$, and can be done via calling the alternating
  quantizer of~\citep{XuYaLiOuCaWaZh18}:
  $B\alpha=\q_{\rm alt}(\wt{\theta})$.
\end{enumerate}
Together, the prox operator generalizes the alternating minimization
procedure in~\citep{XuYaLiOuCaWaZh18}, as $\lambda$ governs a
trade-off between quantization and closeness to $\theta$.  To see that
this is a strict generalization, note that for any $\lambda$ the
solution of~\cref{equation:multi-bit-prox} will be an interpolation
between the input $\theta$ and its Euclidean projection to
$\mc{Q}$. As $\lambda \rightarrow +\infty$, the prox operator
collapses to the projection.

\paragraph{Ternary quantization}
Ternary quantization is a variant of 2-bit quantization, in which
weights are constrained to be in $\set{-\alpha,0,\beta}$ for real
values $\alpha,\beta>0$. We defer the derivation of the ternary prox
operator into Appendix~\ref{appendix:ternary-prox}.



\subsection{Homotopy method for regularization strength}
\label{section:homotopy}
Recall that the larger $\lambda_t$ is, the more aggressive
$\theta_{t+1}$ will move towards the quantized set. 
An ideal choice would be to (1) force the net to be exactly quantized
upon convergence, and (2) not be too aggressive such that the
quantized net at convergence is sub-optimal.

We let $\lambda_t$ be a linearly increasing sequence,
i.e. $\lambda_t\defeq \lambda\cdot t$ for some hyper-parameter
$\lambda>0$ which we term as the \emph{regularization rate}. With this
choice, the stochastic gradient steps will start off close to
full-precision training and gradually move towards exact
quantizedness, hence the name ``homotopy method''.  The parameter
$\lambda$ can be tuned by minimizing the validation loss, and controls
the aggressiveness of falling onto the quantization constraint. There
is nothing special about the linear increasing scheme, but it is
simple enough and works well as we shall see in the experiments.

%% file: Sections/experiment-new.tex
\section{Experiments}
\label{section:experiment}
We evaluate the performance of \pq~on two tasks: image classification
with ResNets, and language modeling with LSTMs. On both tasks, we show
that the default straight-through gradient method is not the only
choice, and our \pq~can achieve the same and often better results.

\subsection{Image classification on CIFAR-10}
\label{section:cifar-10}
\paragraph{Problem setup} We perform image classification on the
CIFAR-10 dataset, which contains 50000 training images and 10000 test
images of size 32x32. We apply a commonly used data augmentation
strategy (pad by 4 pixels on each side, randomly crop to 32x32, do a
horizontal flip with probability 0.5, and normalize). Our models are
ResNets~\citep{HeZhReSu16} of depth 20, 32, 44, and 56 with weights
quantized to binary or ternary.

\paragraph{Method} We use \pq~with
regularizer~\cref{equation:binary-reg} in the binary case
and~\cref{equation:ternary-prox,equation:ternary-quantizer} in the
ternary case, which we respectively denote as PQ-B and PQ-T.  We use
the homotopy method $\lambda_t=\lambda\cdot t$ with $\lambda=10^{-4}$
as the regularization strength and Adam with constant learning rate
0.01 as the optimizer.

We compare with BinaryConnect (BC) for binary nets and Trained Ternary
Quantization (TTQ)~\citep{ZhuHaMaDa16} for ternary nets. For
BinaryConnect, we train with the recommended Adam optimizer with
learning rate decay~\citep{CourbariauxBeDa15} (initial learning rate
0.01, multiply by 0.1 at epoch 81 and 122), which we find leads to the
best result for BinaryConnect. For TTQ we compare with the reported
results in~\citep{ZhuHaMaDa16}.

For binary quantization, both BC and our \pq~are initialized at the
same pre-trained full-precision nets (warm-start) and trained for 300
epochs for fair comparison.  For both methods, we perform a hard
quantization $\theta\mapsto\q(\theta)$ at epoch 200 and keeps training
till the 300-th epoch to stabilize the BatchNorm layers. We compare in
addition the performance drop relative to full precision nets of
BinaryConnect, BinaryRelax~\citep{YinZhLyOsQiXi18}, and our \pq.

\paragraph{Result} The top-1 classification errors for binary
quantization are reported in Table~\ref{table:cifar-10}. Our
\pq~consistently yields better results than BinaryConnect. The
performance drop of \pq~relative to full-precision nets is about
$1\%$, better than BinaryConnect by
$0.2\%$ on average and significantly better than the
reported result of BinaryRelax.

Results and additional details
for ternary quantization are deferred to
Appendix~\ref{appendix:ternary-experiment}.

\begin{table}[h!]
  \centering
  \caption{Top-1 classification error of binarized ResNets on
    CIFAR-10. Performance is reported in mean(std) over 4 runs, as well as the (absolute) performance drop of over full-precision nets.}
  \label{table:cifar-10}
  \begin{tabular}{|c|c||c|c||c|c|c|}
    \hline
    \multicolumn{2}{|c||}{} & \multicolumn{2}{|c||}{Classification error} & \multicolumn{3}{|c|}{Performance drop over FP net} \\
    \hline
    Model & FP & BC & PQ-B (ours) & BC & BinaryRelax & PQ-B (ours) \\  
    (Bits) & (32) & (1) & (1) & (1) & (1) & (1)\\
    \hline
    ResNet-20 & 8.06 & 9.54 (0.03) & {\bf 9.35} (0.13) & +1.48 & +4.84 & {\bf +1.29} \\ 
    ResNet-32 & 7.25 & 8.61 (0.27) & {\bf 8.53} (0.15) & +1.36 & +2.75 & {\bf +1.28} \\
    ResNet-44 & 6.96 & 8.23 (0.23) & {\bf 7.95} (0.05) & +1.27 & - & {\bf +0.99} \\
    ResNet-56 & 6.54 & 7.97 (0.22) & {\bf 7.70} (0.06) & +1.43 & - & {\bf +1.16} \\
    \hline
  \end{tabular}
\end{table}


\subsection{Language modeling with LSTMs}
\label{section:lstm}
\paragraph{Problem setup} We perform language modeling with
LSTMs~\cite{HochreiterSc97} on the Penn Treebank (PTB)
dataset~\citep{MarcusMaSa93}, which contains 929K training tokens, 73K
validation tokens, and 82K test tokens. Our model is a standard
one-hidden-layer LSTM with embedding dimension 300 and hidden
dimension 300. We train quantized LSTMs with the encoder, transition
matrix, and the decoder quantized to $k$-bits for
$k\in\set{1,2,3}$. The quantization is performed in a row-wise
fashion, so that each row of the matrix has its own codebook
$\set{\alpha_1,\dots,\alpha_k}$.

\paragraph{Method}
We compare our multi-bit \pq~(\cref{equation:multi-bit-prox}) to the
state-of-the-art alternating minimization algorithm with
straight-through gradients~\citep{XuYaLiOuCaWaZh18}.
Training is initialized at a
pre-trained full-precision LSTM. We use the SGD optimizer with initial
learning rate 20.0 and decay by a factor of 1.2 when the validation
error does not improve over an epoch. We train for 80 epochs with
batch size 20, BPTT 30, dropout with probability 0.5, and clip the
gradient norms to $0.25$. The regularization rate $\lambda$ is tuned
by finding the best performance on the validation set. In addition to
multi-bit quantization, we also report the results for binary LSTMs
(weights in $\set{\pm 1}$), comparing BinaryConnect and our
\pq -Binary, where both learning rates are tuned on an exponential
grid $\set{2.5, 5, 10, 20, 40}$.


\paragraph{Result}
We report the perplexity-per-word (PPW, lower is better) in
Table~\ref{table:ptb}. The performance of \pq~is comparable with the
Straight-through gradient method. On Binary LSTMs, \pq-Binary beats
BinaryConnect by a large margin. These results demonstrate that
\pq~offers a powerful alternative for training recurrent networks.

\begin{table}[h!]
  \centering
  \caption{PPW of quantized LSTM on Penn Treebank.}
  \label{table:ptb}
  \begin{tabular}{|c|c|c|c|c|}
    \hline
    Method / Number of Bits & 1 & 2 & 3 & FP (32) \\
    \hline
    BinaryConnect & 372.2 & - & - & \multirow{4}{*}{88.5} \\
    \cline{1-4}
    \pq-Binary (ours) & {\bf 288.5} & - & - & \\
    \cline{1-4}
    ALT Straight-through\footnotemark & 104.7 & 90.2 & 86.1 & \\
    \cline{1-4}
    ALT-\pq~(ours) & 106.2 & 90.0 & 87.2 & \\
    \hline
  \end{tabular}
\end{table}

\footnotetext{We thank~\citet{XuYaLiOuCaWaZh18} for sharing the
  implementation of this method through a personal
  communication. There is a very clever trick not mentioned in their
  paper: after computing the alternating quantization $\q_{\rm alt}(\theta)$,
  they multiply by a constant 0.3 before taking the gradient; in other
  words, their quantizer is a rescaled alternating quantizer:
  $\theta\mapsto 0.3\q_{\rm alt}(\theta)$. This scaling step gives a
  significant gain in performance -- without scaling the PPW is
  $\set{116.7, 94.3, 87.3}$ for $\set{1,2,3}$ bits. In contrast, our
  \pq~does not involve a scaling step and achieves better PPW than
  this unscaled ALT straight-through method.}


%% file: Sections/theory.tex
\section{Theoretical analysis}
\label{section:theory}
In this section, we perform a theoretical study on the convergence of
quantization algorithms. We show in
Section~\ref{section:convergence-pq} that our \pq~algorithm
(i.e. non-lazy prox-gradient method) converges under mild smoothness
assumptions on the problem. In
Section~\ref{section:nonconvergence-lazy}, we provide a simple example
showing that the lazy prox-gradient method fails to converge under the
same set of assumptions. In Section~\ref{section:convergence-bc}, we
show that BinaryConnect has a very stringent condition for converging
to a fixed point.  Our theory demonstrates the superiority of our
proposed \pq~over lazy prox-gradient type algorithms such as
BinaryConnect and BinaryRelax~\citep{YinZhLyOsQiXi18}. All missing
proofs are deferred to Appendix~\ref{appendix:proof}.

Prox-gradient algorithms (both lazy and non-lazy) with a
fixed $\lambda$ aim to solve the problem
\begin{equation}
  \label{problem:regularized}
  \minimize_{\theta\in\R^d} L(\theta) + \lambda R(\theta),
\end{equation}
and BinaryConnect can be seen as the limiting case of the above with
$\lambda=\infty$ (cf. Section~\ref{section:lazy-projection}).

\subsection{A convergence theorem for \pq}
\label{section:convergence-pq}
We consider \pq~with batch gradient and constant regularization strength $\lambda_t\equiv \lambda$:
\begin{equation*}
  \theta_{t+1} = \prox_{\eta_t\lambda R}(\theta_t - \eta_t\grad L(\theta_t)).
\end{equation*}


\begin{theorem}[Convergence of ProxQuant]
  \label{theorem:convergence-pq}
  Assume that the loss $L$ is $\beta$-smooth (i.e. has
  $\beta$-Lipschitz gradients) and the regularizer $R$ is
  differentiable. Let $F_\lambda(\theta)=L(\theta)+\lambda R(\theta)$
  be the composite objective and assume that it is bounded below by
  $F_\star$. Running ProxQuant with batch gradient $\grad L$, constant
  stepsize $\eta_t \equiv \eta = \frac{1}{2\beta}$ and
  $\lambda_t\equiv \lambda$ for $T$ steps, we have the convergence
  guarantee
  \begin{equation}
    \label{equation:convergence-pq}
    \ltwo{\grad F_\lambda(\theta_{T_{\rm best}})}^2 \le
    \frac{C\beta(F_\lambda(\theta_0) - F_\star)}{T}~~~{\rm
      where}~~~T_{\rm best} = \argmin_{1\le t\le T} \ltwo{\theta_t - \theta_{t-1}},
  \end{equation}
  where $C>0$ is a universal constant.
\end{theorem}

\begin{remark}
  The convergence guarantee requires both the loss and the regularizer
  to be smooth. Smoothness of the loss can be satisfied if we use a
  smooth activation function (such as $\tanh$). For the regularizer,
  the quantization-inducing regularizers defined in
  Section~\ref{section:wasserstein-reg} (such as the W-shaped
  regularizer) are non-differentiable.
  However,
  we can use a smoothed version of them that is differentiable and
  point-wise arbitrarily close to $R$, which will satisfy the
  assumptions of Theorem~\ref{theorem:convergence-pq}. The proof of Theorem~\ref{theorem:convergence-pq} is deferred to Appendix~\ref{appendix:proof-convergence-pq}.
\end{remark}

\subsection{Non-convergence of lazy prox-gradient}
\label{section:nonconvergence-lazy}
The lazy prox-gradient algorithm (e.g. BinaryRelax~\citep{YinZhLyOsQiXi18}) for solving problem~\cref{problem:regularized} is a variant where the gradients are taken at proximal points but accumulated at the original sequence:
\begin{equation}
  \label{algorithm:lazy}
  \theta_{t+1} = \theta_t - \eta_t\grad L(\prox_{\lambda R}(\theta_t)).
\end{equation}
Convergence of the lazy prox-gradient algorithm~\cref{algorithm:lazy}
is only known to hold for convex problems~\citep{Xiao10}; on smooth
non-convex problems it generally does not converge even in an ergodic
sense. We provide a concrete example that satisfies the
assumptions in Theorem~\ref{theorem:convergence-pq} (so that
\pq~converges ergodically) but lazy prox-gradient does not converge.

\begin{theorem}[Non-convergence of lazy prox-gradient]
  \label{theorem:nonconvergence-lazy}
  There exists $L$ and $R$ satisfying the assumptions of
  Theorem~\ref{theorem:convergence-pq} such that for any constant
  stepsize $\eta_t\equiv \eta\le \frac{1}{2\beta}$, there exists some
  specific initialization $\theta_0$ on which the lazy prox-gradient
  algorithm~\cref{algorithm:lazy} oscillates between two
  non-stataionry points and hence does not converge in the ergodic
  sense of~\cref{equation:convergence-pq}.
\end{theorem}

\begin{remark}
  Our construction is a fairly simple example in one-dimension and not very adversarial: $L(\theta)=\frac{1}{2}\theta^2$ and $R$ is a smoothed W-shaped regularizer. See Appendix~\ref{appendix:proof-nonconvergence-lazy} for the details.
\end{remark}

\subsection{Convergence characterization for BinaryConnect}
\label{section:convergence-bc}
For BinaryConnect, the concept of stataionry points is no longer
sensible (as the target points $\set{\pm 1}^d$ are isolated and hence
every point is stationary). Here, we consider the alternative
definition of convergence as converging to a fixed point and show that
BinaryConnect has a very stringent convergence condition.

Consider the BinaryConnect method with batch gradients:
\begin{equation}
  \label{equation:batch-bc}
  \begin{aligned}
    s_t = \sign(\theta_t),~~~\theta_{t+1} = \theta_t - \eta_t\grad L(s_t).
  \end{aligned}
\end{equation}

\begin{definition}[Fixed point and convergence]
  We say that $s\in\{\pm 1\}^d$ is a \textbf{fixed point} of the
  BinaryConnect algorithm, if $s_0=s$ in~\cref{equation:batch-bc}
  implies that $s_t = s$ for all $t = 1,2,...$.  We say that the
  BinaryConnect algorithm \textbf{converges} if there exists
  $t < \infty$ such that $s_t$ is a fixed point.
\end{definition}
\begin{theorem}
  \label{theorem:fixed-point}
  Assume that the learning rates satisfy
  $\sum_{t=0}^\infty \eta_t = \infty$, then
  $s\in \{\pm 1\}^d$ is a fixed point for
  BinaryConnect~\cref{equation:batch-bc} \emph{if and only if}
  $\sign(\nabla L(s)[i])=-s[i]$ for all $i\in[d]$ such that
  $\nabla L(\theta)[i]\neq 0$. Such a point may not exist, in which
  case BinaryConnect does not converge for any initialization
  $\theta_0\in \R^d$.
\end{theorem}

\begin{remark}
  Theorem~\ref{theorem:fixed-point} is in appearingly a stark contrast
  with the convergence result for BinaryConnect
  in~\citep{LiDeXuStSaGo17} in the convex case, whose bound involves a
  an additive error $O(\Delta)$ that does not vanish over iterations,
  where $\Delta$ is the grid size for quantization. Hence, their
  result is only useful when $\Delta$ is small. In contrast, we
  consider the original BinaryConnect with $\Delta = 1$, in which case
  the error makes ~\citet{LiDeXuStSaGo17}'s bound vacuous. The proof
  of Theorem~\ref{theorem:fixed-point} is deferred to
  Appendix~\ref{appendix:proof-fixed-point}.
\end{remark}

\paragraph{Experimental evidence}
We have already seen that such a fixed point $s$ might not exist in
the toy example in Figure~\ref{figure:bc-sucks}. In Appendix~\ref{appendix:sign-change},
we perform a sign change experiment on CIFAR-10, showing that
BinaryConnect indeed fails to converge to a fixed sign pattern,
corroborating Theorem~\ref{theorem:fixed-point}.



%% file: Sections/conclusion.tex
\section{Conclusion}
In this paper, we propose and experiment with the \pq~method for
training quantized networks. Our results demonstrate that \pq~offers a
powerful alternative to the straight-through gradient method and
has theoretically better convergence properties.
For future work, it would
be of interest to propose alternative regularizers for ternary and
multi-bit \pq~and experiment with our method on larger tasks.

%% file: Sections/acknowledgement.tex
\section*{Acknowledgement}
We thank Tong He, Yifei Ma, Zachary Lipton, and John Duchi for their valuable feedback. We thank Chen Xu and Zhouchen Lin for the insightful discussion on multi-bit quantization and sharing the implementation of~\citep{XuYaLiOuCaWaZh18} with us. We thank Ju Sun for sharing the draft of~\citep{Sun18} and the inspiring discussions on adversarial regularization for quantization.
The majority of this work was performed when YB and YW were at Amazon AI. 

%% file: Sections/wasserstein.tex
\section{Additional results on Regularization}

\subsection{Prox operators for binary nets}
\label{appendix:prox}
Here we derive the prox operators for the binary
regularizer~\cref{equation:w-binary} and its squared $L_2$
variant. Recall that
\begin{equation*}
  R_{\rm bin}(\theta) = \sum_{j=1}^d \min\set{|\theta_j-1|, |\theta_j+1|}.
\end{equation*}
By definition of the prox operator, we have for any $\theta\in\R^d$
that
\begin{align*}
  & \quad \prox_{\lambda R_{\rm bin}}(\theta) = \argmin_{\wt{\theta}\in\R^d}
    \set{\frac{1}{2}\ltwo{\wt{\theta}-\theta}^2 + \lambda\sum_{j=1}^d
    \min\set{|\wt{\theta}_j-1|, |\wt{\theta}_j+1|}} \\ 
  & = \argmin_{\wt{\theta}\in\R^d} \set{ \sum_{j=1}^d \frac{1}{2}(\wt{\theta}_j -
    \theta_j)^2 + \lambda\min\set{|\wt{\theta}_j-1|, |\wt{\theta}_j+1|}}.
\end{align*}
This minimization problem is coordinate-wise separable. For each
$\wt{\theta}_j$, the penalty term remains the same upon flipping the sign,
but the quadratic term is smaller when
$\sign(\wt{\theta}_j)=\sign(\theta_j)$. Hence, the solution
$\theta^\star$ to the prox satisfies that
$\sign(\theta^\star_j)=\sign(\theta_j)$, and the absolute value
satisfies
\begin{equation*}
  |\theta_j^\star| = \argmin_{t\ge 0} \set{\frac{1}{2}(t -
    |\theta_j|)^2 + \lambda|t-1|} = \st(|\theta_j|, 1, \lambda) =
  1 + \sign(|\theta_j| - 1)[||\theta_j| - 1| - \lambda]_+.
\end{equation*}
Multiplying by $\sign(\theta_j^\star)=\sign(\theta_j)$, we have
\begin{equation*}
  \theta_j^\star = \st(\theta_j, \sign(\theta_j), \lambda),
\end{equation*}
which gives~\cref{equation:binary-prox}.

For the squared $L_2$ version, by a similar argument, the
corresponding regularizer is
\begin{equation*}
  R_{\rm bin}(\theta) = \sum_{j=1}^d \min\set{(\theta_j-1)^2, (\theta_j+1)^2}. 
\end{equation*}
For this regularizer we have
\begin{align*}
\prox_{\lambda R_{\rm bin}}(\theta) = \argmin_{\wt{\theta}\in\R^d} \set{
  \sum_{j=1}^d \frac{1}{2}(\wt{\theta}_j - \theta_j)^2 +
  \lambda\min\set{(\wt{\theta}_j-1)^2, (\wt{\theta}_j+1)^2}}.
\end{align*}
Using the same argument as in the $L_1$ case, the solution
$\theta^\star$ satisfies $\sign(\theta^\star_j)=\sign(\theta_j)$,
and
\begin{equation*}
  |\theta_j^\star| = \argmin_{t\ge 0} \set{ \frac{1}{2}(t -
    |\theta_j|)^2 + \lambda(t - 1)^2 } = \frac{|\theta_j| +
    \lambda}{1+\lambda}. 
\end{equation*}
Multiplying by $\sign(\theta_j^\star)=\sign(\theta_j)$ gives
\begin{equation*}
  \theta_j^\star =
  \frac{\theta_j+\lambda\sign(\theta_j)}{1+\lambda},
\end{equation*}
or, in vector form,
$\theta^\star=(\theta+\lambda\sign(\theta))/(1+\lambda)$.

\subsection{Prox operator for ternary quantization}
\label{appendix:ternary-prox}
For ternary quantization, we use an approximate version of the
alternating prox operator~\cref{equation:multi-bit-prox}: compute
$\wt{\theta}=\prox_{\lambda R}(\theta)$ by initializing at 
$\wt{\theta}=\theta$ and repeating
\begin{equation}
  \label{equation:ternary-prox}
  \what{\theta} = \q(\wt{\theta})~~~{\rm and}~~~\wt{\theta} = \frac{\theta + 2\lambda\what{\theta}}{1+2\lambda},
\end{equation}
where $\q$ is the ternary quantizer defined as
\begin{equation}
  \label{equation:ternary-quantizer}
  \q(\theta) = \theta^+\ones\{\theta \ge \Delta\} +
  \theta^-\ones\{\theta \le -\Delta\},~~\Delta =
  \frac{0.7}{d}\lone{\theta},~~\theta^+ =
  \overline{\theta|_{i:\theta_i\ge\Delta}},~~\theta^- =
  \overline{\theta|_{i:\theta_i\le -\Delta}}.
\end{equation}
This is a straightforward extension of the TWN
quantizer~\citep{LiLi16} that allows different levels for positives
and negatives. We find that two rounds of alternating computation
in~\cref{equation:ternary-prox} achieves a good performance,
which we use in our experiments.

%% file: Sections/appendix-experiment.tex
\section{Additional experimental results}
\subsection{Ternary quantization for CIFAR-10}
\label{appendix:ternary-experiment}
Our models are ResNets of depth 20, 32, and 44. Ternarized training
is initialized at pre-trained full-precision nets. We perform a hard
quantization $\theta\mapsto\q(\theta)$ at epoch 400 and
keeps training till the600-th epoch to stabilize the BatchNorm
layers.

\paragraph{Result}
The top-1 classification errors for ternary
quantization are reported in Table~\ref{table:cifar-10-ternary}.
Our results are comparable with the reported results of
TTQ,\footnote{We note that our \pq-Ternary and TTQ are not strictly
  comparable: we have the advantage of using better initializations;
  TTQ has the advantage of a stronger quantizer: they train the
  quantization levels $(\theta^+,\theta^-)$ whereas our
  quantizer~\cref{equation:ternary-quantizer} pre-computes them from
  the current full-precision parameter.} and the best performance of
our method over 4 runs (from the same initialization) is slightly
better than TTQ.

\begin{table}[h!]
  \centering
  \caption{Top-1 classification error of ternarized ResNets on
    CIFAR-10. Performance is reported in mean(std) over 4 runs, where
    for PQ-T we report in addition the best of 4 (Bo4).}
  \label{table:cifar-10-ternary}
  \begin{tabular}{|c|c||c|c|c|}
    \hline
    Model & FP & TTQ & PQ-T (ours) & PQ-T (ours, Bo4) \\  
    (Bits) & (32) & (2) & (2) & (2) \\
    \hline
    ResNet-20 & 8.06 & 8.87 & {\bf 8.40} (0.13) & {\bf 8.22} \\ 
    ResNet-32 & 7.25 & 7.63 & 7.65 (0.15) & {\bf 7.53} \\
    ResNet-44 & 6.96 & 7.02 & 7.05 (0.08) & {\bf 6.98} \\
    \hline
  \end{tabular}
\end{table}

%% file: Sections/signchange.tex
\section{Sign change experiment}
\label{appendix:sign-change}
We experimentally compare the training dynamics of \pq-Binary and
BinaryConnect through the \emph{sign change} metric.
The sign change metric between any $\theta_1$ and $\theta_2$ is the
proportion of their different signs, i.e. the (rescaled) Hamming
distance:
\begin{equation*}
  \signch(\theta_1, \theta_2) = 
  \frac{\lone{\sign(\theta_1) - \sign(\theta_2)}}{2d} \in [0,1].
\end{equation*}
In $\R^d$, the space of all full-precision parameters, the sign
change is a natural distance metric that represents the closeness of
the binarization of two parameters.

\begin{figure}[h!]
  \centering
  \begin{subfigure}[b]{0.23\textwidth}
    \includegraphics[width=\textwidth]{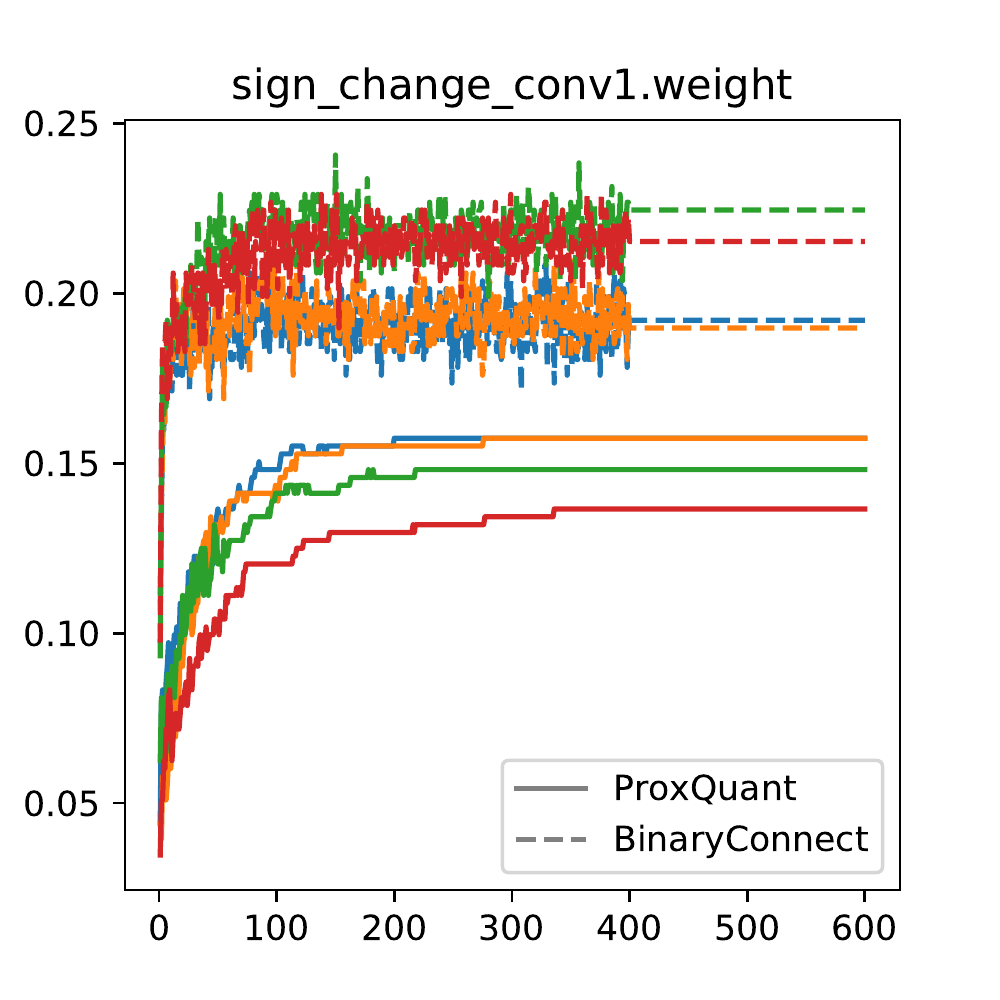}
    \caption{}
  \end{subfigure}
  ~
  \begin{subfigure}[b]{0.23\textwidth}
    \includegraphics[width=\textwidth]{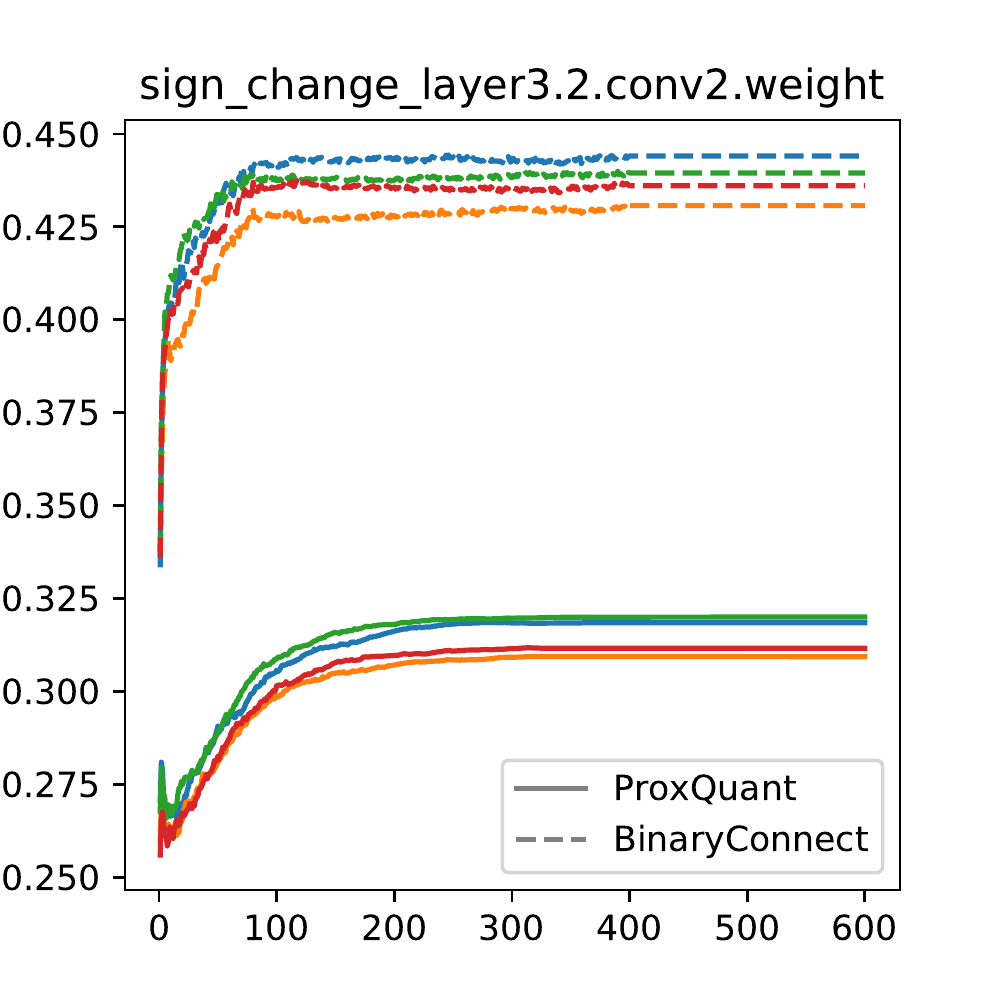}
    \caption{}
  \end{subfigure}
  ~
  \begin{subfigure}[b]{0.23\textwidth}
    \includegraphics[width=\textwidth]{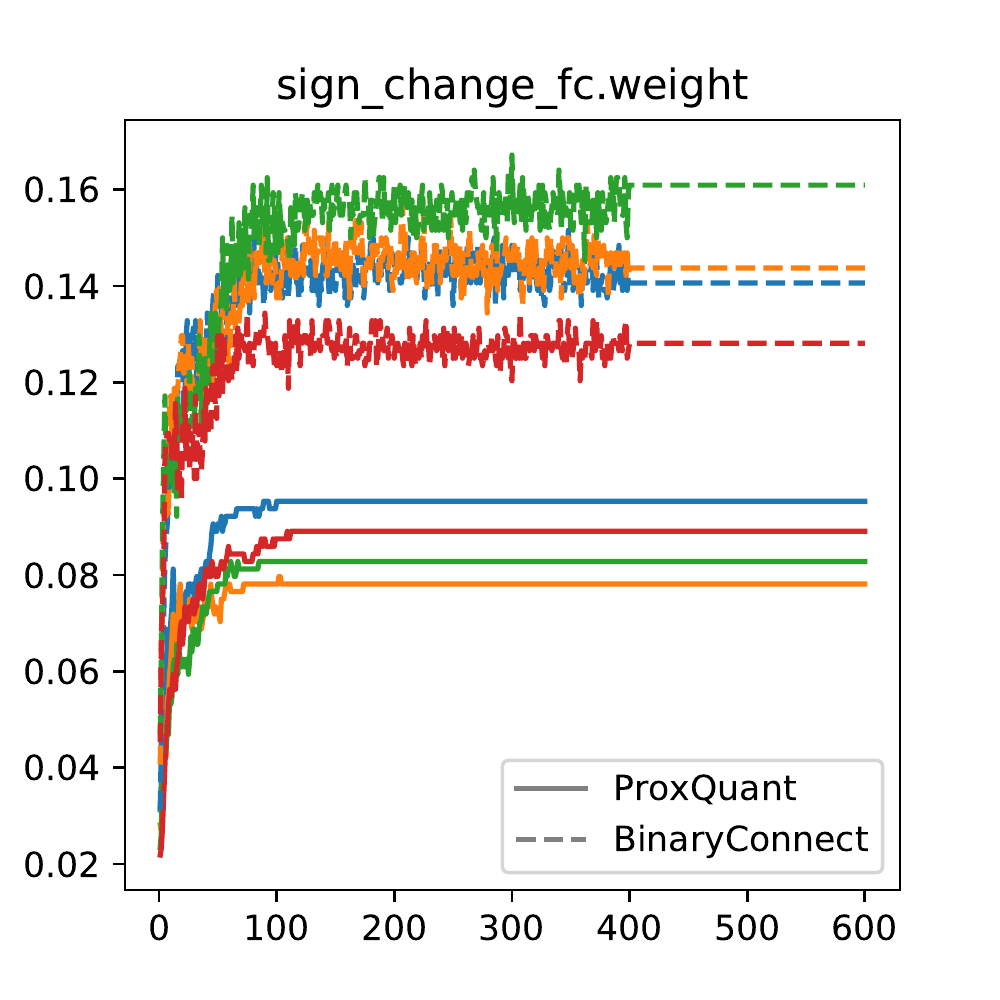}
    \caption{}
  \end{subfigure}
  ~
  \begin{subfigure}[b]{0.23\textwidth}
    \includegraphics[width=\textwidth]{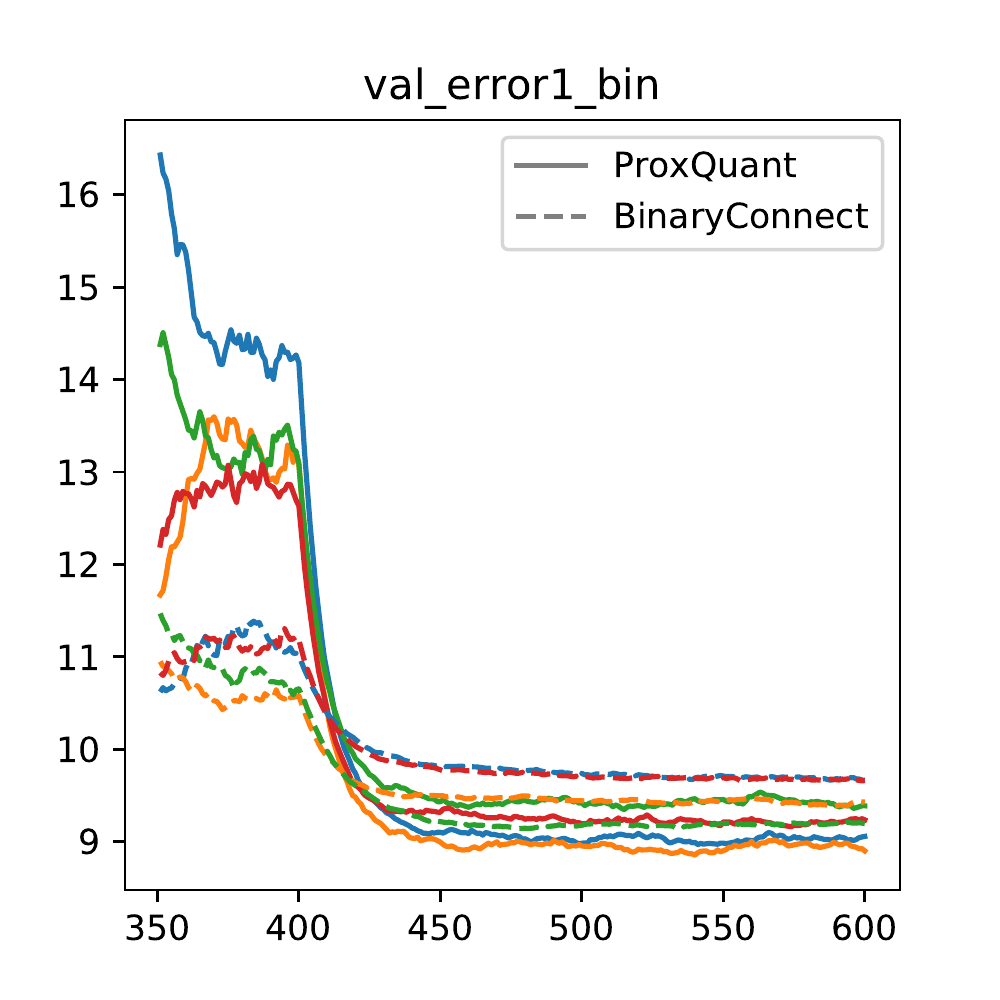}
    \caption{}
  \end{subfigure}
  \caption{\small $\signch(\theta_0,\theta_t)$ against $t$ (epoch) for
    BinaryConnect and \pq, over 4 runs starting from the same
    full-precision ResNet-20.
    \pq~has significantly lower sign changes than BinaryConnect while
    converging to better models.
    (a) The first conv layer of size
    $16\times3\times3\times3$;
    (b) The last conv layer of size
    $64\times64\times3\times3$;
    (c) The fully connected layer of size $64\times10$;
    (d) The validation top-1 error of the binarized nets (with moving
    average smoothing). }
  \label{figure:sign-change}
\end{figure}

Recall in our CIFAR-10 experiments
(Section~\ref{section:cifar-10}), for both BinaryConnect and \pq, we
initialize at a good full-precision net $\theta_0$ and stop at a
converged binary network $\what{\theta}\in\set{\pm 1}^d$.
We are interested
in $\signch(\theta_0,\theta_t)$ along the training path, as well as
$\signch(\theta_0,\what{\theta})$, i.e. the distance of the final
output model to the initialization.


Our finding is that
\pq~produces binary nets with both \emph{lower} sign changes and
\emph{higher} performances, compared with BinaryConnect. Put
differently, around the warm start, there is a good binary net nearby
which can be found by \pq~but not BinaryConnect, suggesting that
BinaryConnect, and in general the straight-through gradient method,
suffers from higher optimization instability than~\pq.
This finding is consistent in all layers,
across different warm starts, and across differnent runs from each
same warm start (see Figure~\ref{figure:sign-change} and
Table~\ref{table:sign-change} in Appendix~\ref{appendix:sign-change-raw}).
This result here is also
consistent with Theorem~\ref{theorem:fixed-point}: the signs in
BinaryConnect never stop changing until we manually freeze the signs
at epoch 400.


\subsection{Raw data for sign change experiment}
\label{appendix:sign-change-raw}
\begin{table}[h]
  \centering
  \caption{Performances and sign changes on ResNet-20 in mean(std)
    over 3 full-precision initializations and 4 runs per
    (initialization x method). Sign changes are computed over all
    quantized parameters in the net.}
  \label{table:sign-change}
  \begin{tabular}{|c|c|c|c|}
    \hline
    Initialization & Method & Top-1 Error(\%) & Sign change \\
    \hline
    FP-Net 1 & BC & 9.489 (0.223) & 0.383 (0.006) \\
    \cline{2-4}
    (8.06) & PQ-B & {\bf 9.146} (0.212) & {\bf 0.276} (0.020) \\
    \hline
    FP-Net 2 & BC & 9.745 (0.422) & 0.381 (0.004)\\
    \cline{2-4}
    (8.31) & PQ-B & {\bf 9.444} (0.067) & {\bf 0.288} (0.002) \\
    \hline
    FP-Net 3 & BC & 9.383 (0.211) & 0.359 (0.001) \\
    \cline{2-4}
    (7.73) & PQ-B & {\bf 9.084} (0.241) & {\bf 0.275} (0.001) \\
    \hline
  \end{tabular}
\end{table}

\begin{table}[h]
  \centering
  \caption{Performances and sign changes on ResNet-20 in raw data over
    3 full-precision initializations and 4 runs per (initialization x
    method). Sign changes are computed over all quantized parameters
    in the net.}
  \label{table:sign-change-raw}
  \begin{tabular}{|c|c|c|c|}
    \hline
    Initialization & Method & Top-1 Error(\%) & Sign change \\
    \hline
    FP-Net 1 & BC & 9.664, 9.430, 9.198, 9.663 & 0.386, 0.377, 0.390,
                                                 0.381 \\
    \cline{2-4}
    (8.06) & PQ-B & 9.058, 8.901, 9.388, 9.237 & 0.288, 0.247, 0.284,
                                                 0.285\\
    \hline
    FP-Net 2 & BC & 9.456, 9.530, 9.623, 10.370 & 0.376, 0.379, 0.382,
    0.386 \\
    \cline{2-4}
    (8.31) & PQ-B & 9.522, 9.474, 9.410, 9.370 & 0.291, 0.287, 0.289,
                                                 0.287 \\
    \hline
    FP-Net 3 & BC & 9.107, 9.558, 9.538, 9.328 & 0.360, 0.357, 0.359,
                                                 0.360 \\
    \cline{2-4}
    (7.73) & PQ-B & 9.284, 8.866, 9.301, 8.884 & 0.275, 0.276, 0.276,
                                                 0.275 \\
    \hline
  \end{tabular}
\end{table}

%% file: Sections/proof.tex
\newpage

\section{Proofs of theoretical results}
\label{appendix:proof}
\subsection{Proof of Theorem~\ref{theorem:convergence-pq}}
\label{appendix:proof-convergence-pq}
Recall that a function $f:\R^d\to\R$ is said to be $\beta$-smooth if it is differentiable and $\grad f$ is $\beta$-Lipschitz: for all $x,y\in\R^d$ we have
\begin{equation*}
  \ltwo{\grad f(x) - \grad f(y)} \le \beta\ltwo{x-y}.
\end{equation*}
For any $\beta$-smooth function, it satisfies the bound
\begin{equation*}
  f(y) \le f(x) + \<\grad f(x), y-x\> + \frac{\beta}{2}\ltwo{x-y}^2~~~{\rm for~all}~x,y\in\R^d.
\end{equation*}

Convergence results like Theorem~\ref{theorem:convergence-pq} are
standard in the literature of proximal algorithms, where we have
convergence to stataionarity without convexity on either $L$ or $R$
but assuming smoothness. For completeness we provide a proof
below. Note that though the convergence is ergodic, the best index
$T_{\rm best}$ can be obtained in practice via monitoring the proximity
$\ltwo{\theta_t - \theta_{t-1}}$.

\begin{proof-of-theorem}[\ref{theorem:convergence-pq}]
  Recall the ProxQuant iterate
  \begin{equation*}
    \theta_{t+1} = \argmin_{\theta\in\R^d} \left\{ L(\theta_t) +
      \<\theta-\theta_t, \grad L(\theta_t)\> +
      \frac{1}{2\eta}\ltwo{\theta - \theta_t}^2 + \lambda R(\theta)
    \right\}.
  \end{equation*}
  By the fact that $\theta_{t+1}$ minimizes the above objective and
  applying the smoothness of $L$, we get that
  \begin{align*}
    & \quad F_\lambda(\theta_t) = L(\theta_t) + \lambda R(\theta_t)
      \ge L(\theta_t) + \<\theta_{t+1}
      - \theta_t, \grad L(\theta_t)\> +
      \frac{1}{2\eta}\ltwo{\theta_{t+1} - \theta_t}^2 + \lambda
      R(\theta_{t+1}) \\
    & \ge L(\theta_{t+1}) + \left( \frac{1}{2\eta} - \frac{\beta}{2}
      \right)\ltwo{\theta_{t+1} - \theta_t}^2 + \lambda
      R(\theta_{t+1}) = F_\lambda(\theta_{t+1}) +
      \frac{\beta}{2}\ltwo{\theta_{t+1} - \theta_t}^2.
  \end{align*}
  Telescoping the above bound for $t=0,\dots,T-1$, we get that
  \begin{equation*}
    \sum_{t=0}^{T-1} \ltwo{\theta_{t+1} - \theta_t}^2 \le
    \frac{2(F_\lambda(\theta_0) - F_\lambda(\theta_T))}{\beta} \le
    \frac{2(F_\lambda(\theta_0) - F_\star)}{\beta}.
  \end{equation*}
  Therefore we have the proximity guarantee
  \begin{equation}
    \label{equation:proximity}
    \min_{0\le t\le T-1} \ltwo{\theta_{t+1} - \theta_t}^2 \le
    \frac{2(F_\lambda(\theta_0) - F_\star)}{\beta T}. 
  \end{equation}
  We now turn this into a stationarity guarantee. The first-order
  optimality condition for $\theta_{t+1}$ gives
  \begin{equation*}
    \grad L(\theta_t) + \frac{1}{\eta}(\theta_{t+1} - \theta_t) +
    \lambda \grad R(\theta_{t+1}) = 0.
  \end{equation*}
  Combining the above equality and the smoothness of $L$, we get
  \begin{align*}
    & \quad \ltwo{\grad F_\lambda(\theta_{t+1})} = \ltwo{\grad L(\theta_{t+1}) +
      \lambda\grad R(\theta_{t+1})} = \ltwo{\frac{1}{\eta}(\theta_t -
      \theta_{t+1}) + \grad L(\theta_{t+1}) - \grad L(\theta_t)} \\
    & \le \left( \frac{1}{\eta} + \beta\right) \ltwo{\theta_{t+1} -
      \theta_t} = 3\beta\ltwo{\theta_{t+1} - \theta_t}.
  \end{align*}
  Choosing $t=T_{\rm best}-1$ and applying the proximity
  guarantee~\cref{equation:proximity}, we get
  \begin{equation*}
    \ltwo{\grad F_\lambda(\theta_{T_{\rm best}})}^2 \le
    9\beta^2 \ltwo{\theta_{T_{\rm best}} - \theta_{T_{\rm best}-1}}^2 =
    9\beta^2 \min_{0\le t\le T-1} \ltwo{\theta_{t+1} - \theta_t}^2 \le
    \frac{18\beta(F_\lambda(\theta_0) - F_\star)}{T}.
  \end{equation*}
  This is the desired bound.
\end{proof-of-theorem}

\subsection{Proof of Theorem~\ref{theorem:nonconvergence-lazy}}
\label{appendix:proof-nonconvergence-lazy}
\begin{wrapfigure}{R}{0.3\textwidth}
  \centering
  \includegraphics[width=0.28\textwidth]{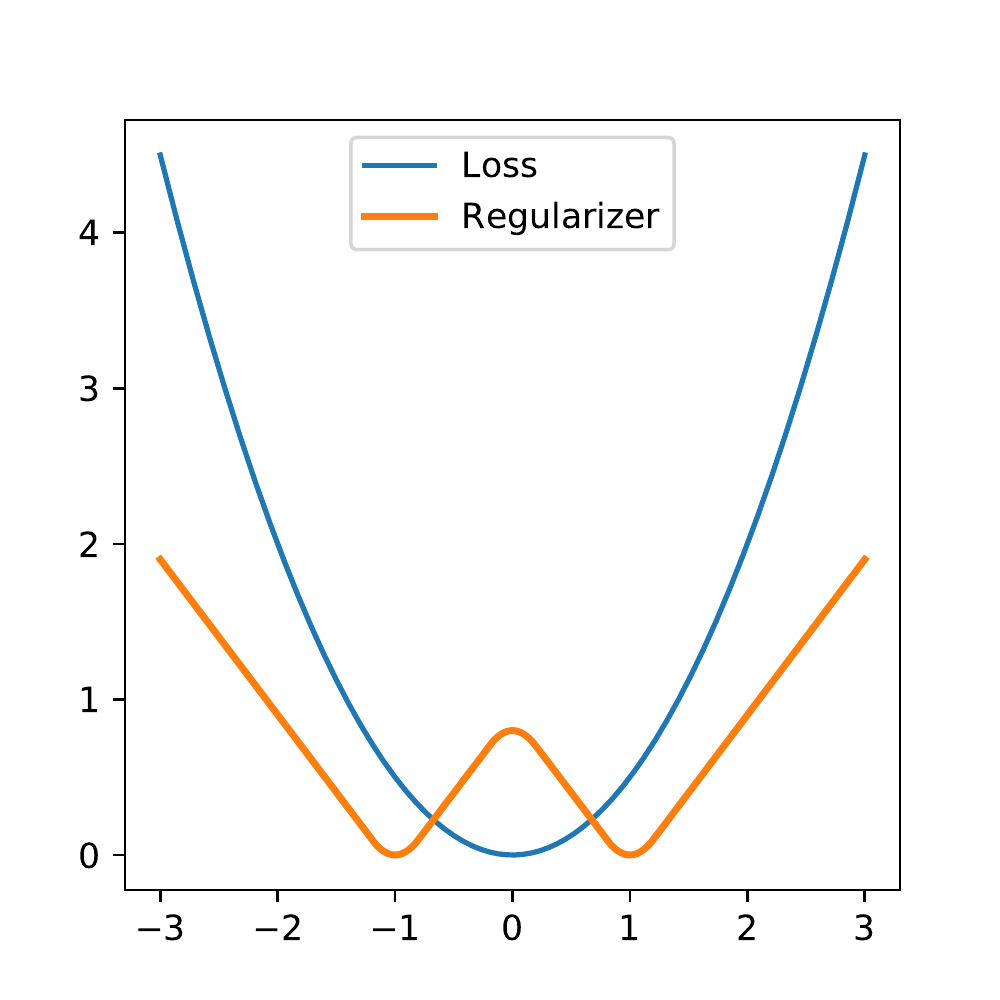}
  \caption{}
  \label{figure:w-smooth}
\end{wrapfigure}

Let our loss function $L:\R\to\R$ be the quadratic $L(\theta)=\frac{1}{2}\theta^2$ (so that $L$ is $\beta$-smooth with $\beta=1$). Let the regularizer $R:\R\to\R$ be a smoothed version of the W-shaped regularizer in~\cref{equation:binary-reg}, defined as (for $\eps\in(0, 1/2]$ being the smoothing radius)

  \begin{equation*}
    R(\theta) = \left\{
      \begin{aligned}
        & -\frac{1}{2\eps}\theta^2 + 1 - \eps, &~~~\theta \in [0, \eps) \\
        & -\theta + 1 - \frac{\eps}{2}, &~~~\theta \in [\eps, 1-\eps) \\
        & \frac{1}{2\eps}(\theta-1)^2, &~~~\theta \in [1-\eps, 1+\eps) \\
        & \theta - 1 - \frac{\eps}{2}, &~~~\theta \in [1+\eps, \infty) \\
      \end{aligned}
    \right.
  \end{equation*}
  and $R(-\theta)=R(\theta)$ for the negative part. See Figure~\ref{figure:w-smooth} for an illustration of the loss $L$ and the regularizer $R$ (with $\eps=0.2$).

  It is straightforward to see that $R$ is piecewise quadratic and differentiable on $\R$ by computing the derivatives at $\eps$ and $1\pm \eps$. Further, by elementary calculus, we can evaluate the prox operator in closed form: for all $\lambda\ge 1$, we have
  \begin{equation*}
    \prox_{\lambda R}(\theta) = \frac{\eps\theta+\lambda\sign(\theta)}{\eps+\lambda\sign(\theta)}~~~\textrm{for all}~|\theta|\le 1.
  \end{equation*}

  Now, suppose we run the lazy prox-gradient method with constant stepsize $\eta_t\equiv \eta\le \frac{1}{2\beta}=\frac{1}{2}$. For the specific initialization
  \begin{equation*}
    \theta_0 = \frac{\eta\lambda}{2\lambda + (2-\eta)\eps} \in (0, 1),
  \end{equation*}
  we have the equality $\prox_{\lambda R}(\theta_0) = \frac{2}{\eta}\theta_0$ and therefore the next lazy prox-gradient iterate is
  \begin{equation*}
    \theta_1 = \theta_0 - \eta\grad L(\prox_{\lambda R}(\theta_0)) = \theta_0 - \eta\grad L\left(\frac{2}{\eta}\theta_0\right) = \theta_0 - \eta\cdot \frac{2}{\eta}\theta_0 = -\theta_0.
  \end{equation*}
  As both $R$ and $L$ are even functions, a symmetric argument holds for $\theta_1$ from which we get $\theta_2=-\theta_1=\theta_0$. Therefore the lazy prox-gradient method ends up oscillating between two points:
  \begin{equation*}
    \theta_t = (-1)^t\theta_0.
  \end{equation*}
  On the other hand, it is straightforward to check that the only stationary points of $L(\theta)+\lambda R(\theta)$ are $0$ and $\pm\frac{\lambda}{\eps+\lambda}$, all not equal to $\pm\theta_0$. Therefore the sequence $\{\theta_t\}_{t\ge 0}$ does not have a subsequence with vanishing gradient and thus does not approach stationarity in the ergodic sense. \qed

\subsection{Proof of Theorem~\ref{theorem:fixed-point}}
\label{appendix:proof-fixed-point}


We start with the ``$\Rightarrow$'' direction.
If $s$ is a fixed point, then by definition there exists $\theta_0\in\R^{d}$ such that $\theta_t = \theta$ for all $t  = 0,1,2,...$. By the iterates \cref{equation:batch-bc}
$$
\theta_T= \theta_0 - \sum_{t=0}^T \eta_t \grad L(s_t).
$$
Take signs on both sides and apply $s_t = s$ for all $t$ on both sides, we get that
$$
s = s_T  =  \sign(\theta_T) =  \sign\left(\theta_0 -   \grad L(s)  \sum_{t=0}^T \eta_t\right)
$$
Take the limit $T\rightarrow \infty$ and apply the assumption that $\sum_t \eta_t = \infty$, we get that for all $i \in [d]$ such that $[\grad L(\theta)]_i\neq 0$,  
$$
s[i]  = \lim_{T\rightarrow \infty}	\sign\left(\theta_0 -   \grad L(s)  \sum_{t=0}^T \eta_t\right)[i] =    - \sign(\grad L(s))[i].
$$

Now we prove the ``$\Leftarrow$'' direction.  If $\theta$ obeys that
$\sign(\nabla L(s)[i])=-s[i]$ for all $i\in[d]$ such that
$\nabla L(s)[i]\neq 0$, then if we take any $\theta_0$ such that
$\sign(\theta_0) = s$, $\theta_t$ will move in a straight line towards the
direction of $-\nabla L(s)$, which does not change the sign of
$h_0$. In other words, $s_t = \sign(\theta_t) = \sign(\theta_0) = s$
for all $t=0,1,2,...$.  Therefore, by definition, $s$ is a fixed
point.